\documentclass[times,final]{elsarticle}
\usepackage{times}
\usepackage{graphicx}
\usepackage{booktabs,multirow}
\usepackage{latexsym}
\usepackage[small]{caption}
\usepackage{subcaption}
\usepackage{graphicx,booktabs,colortbl,multirow}
\usepackage[table,x11names,dvipsnames,table]{xcolor}
\usepackage{listings}
\usepackage{comment}
\usepackage[hyphens]{url}
\usepackage{hyperref}
\usepackage[hyphenbreaks]{breakurl}
\hypersetup{
           breaklinks=true,   
           colorlinks=true   
        }

\usepackage{color} 

\begin{document}
\makeatletter
\def\ps@pprintTitle{%
  \let\@oddhead\@empty
  \let\@evenhead\@empty
  \let\@oddfoot\@empty
  \let\@evenfoot\@oddfoot
}
\makeatother

\begin{frontmatter}

\title{Semi-Supervised Learning for Image Classification using Compact Networks in the BioMedical Context}

\author{A. Inés$^{1}$\corref{mycorrespondingauthor}}\ead{adrian.ines@unirioja.es}
\author{A. Díaz-Pinto$^{2}$, C. Domínguez$^{1}$, J. Heras$^{1}$, E. Mata$^{1}$ and V. Pascual$^{1}$}

\address{$^{1}$University of La Rioja, Department of Mathematics and Computer Science, Spain\\
$^{2}$Department of Biomedical Engineering, School of Biomedical Engineering Imaging Sciences, King’s Collegue London. UK.}

\cortext[mycorrespondingauthor]{Corresponding author}

%
%
%
%
%
%
\begin{abstract}

\textbf{Background and objectives: }The development of mobile and on the edge applications that embed deep convolutional neural models has the potential to revolutionise biomedicine. However, most deep learning models require computational resources that are not available in smartphones or edge devices; an issue that can be faced by means of compact models. The problem with such models is that they are, at least usually, less accurate than bigger models. In this work, we study how this limitation can be addressed with the application of semi-supervised learning techniques.

\textbf{Methods: } We conduct several statistical analyses to compare performance of deep compact architectures when trained using semi-supervised learning methods for tackling image classification tasks in the biomedical context. In particular, we explore three families of compact networks, and two families of semi-supervised learning techniques for 10 biomedical tasks.


\textbf{Results: } By combining semi-supervised learning methods with compact networks, it is possible to obtain a similar performance to standard size networks. In general, the best results are obtained when combining data distillation with MixNet, and plain distillation with ResNet-18. Also, in general, NAS networks obtain better results than manually designed networks and quantized networks.

\textbf{Conclusions: }The work presented in this paper shows the benefits of apply semi-supervised methods to compact networks; this allow us to create compact models that are not only as accurate as standard size models, but also faster and lighter. Finally, we have developed a library that simplifies the construction of compact models using semi-supervised learning methods.


\end{abstract}

\begin{keyword}
Semi-supervised learning, Compact networks, Image Classification, BioMedical Imaging.
\end{keyword}
\end{frontmatter}
\section{Introduction}

Deep learning techniques, and namely convolutional neural networks, have become the state-of-the-art approach to solve image classification problems in biomedicine~\cite{Meijering20}. These techniques have the potential to revolutionise healthcare~\cite{Hinton18}, provided that they become accessible to various populations, a challenge that can be tackled thanks to smartphones or edge devices~\cite{mhealth}. In this scenario, deep learning models need to be executed on edge devices to reduce latency and preserve privacy. Nevertheless, traditional deep learning models have millions of parameters~\cite{efficientnet} and are known to be expensive in terms of computation, memory, and power consumption. These drawbacks make it challenging to embed them in mobile applications~\cite{He16}, a problem that can be faced by using compact deep networks, also known as hardware-aware networks. 

Compact neural networks are designed taking into consideration not only the accuracy of the networks, but also their computational complexity. Initially, these networks were manually designed by pruning bigger networks~\cite{cai2020once}, building networks based on operations that are cost-friendly~\cite{Mobilenetv2}, or applying a number of network compression techniques~\cite{Wiedemann20}. Those manual methods have been recently replaced by neural architecture search (NAS) techniques that automatically search for the most accurate and efficient architecture under memory and space constraints~\cite{FBNet}. In addition, the quantization of existing standard-size models, a technique that converts a deep learning model’s weights to a lower precision such that it needs less computation, has emerged as an alternative to the construction of new compact models~\cite{quantization}.

In the biomedical context, compact deep networks have been successfully employed for glaucoma detection~\cite{Li20}, diabetic retinopathy diagnosis~\cite{Suriyal18} or skin cancer classification~\cite{Chaturvedi20}. Those models are usually trained by applying transfer learning~\cite{Razavian14} (a technique that reuses a model trained in a source task, usually classifying natural images from the ImageNet challenge~\cite{ImageNet}, in a new target task) and generally using manually designed compact networks such as ResNet-18~\cite{Li20} or MobileNet~\cite{Chaturvedi20}. However, automatically designed and quantized compact models are scarce in this context; probably, due to the fact that they are optimised for natural images from the ImageNet challenge; and, it is not clear whether these models can be properly transferred to biomedical images, or whether they obtain better results than manually designed compact networks. In this work, we try to shed light to those questions by conducting a thorough study of several manually designed, automatically designed and quantized compact networks when working with biomedical datasets.



Another important issue of compact models in the biomedical context is that they usually obtain worse results than standard size networks~\cite{Mobilenetv2,FBNet}. In this work, we propose to face this drawback by using semi-supervised learning techniques~\cite{Zhu09}; a set of techniques that employ both labelled and unlabelled data for training a model. Semi-supervised learning might be especially useful in the biomedical context where annotations are scarce and producing them is a time-consuming task that require expert knowledge~\cite{CheXpert}. Semi-supervised learning methods have been employed to improve several models for, among other purposes, classifying chest x-ray abnormalities~\cite{Ho20}, grading diabetic retinopathy~\cite{Luo20}, or analysing gastric x-ray images~\cite{Li20a}. However, semi-supervised learning methods are usually applied with big networks, and their impact on compact networks is not usually explored. Hence, in this paper, we study how compact networks can be combined with semi-supervised learning methods to obtain accurate and efficient models.

Namely, the contributions of this paper are:
\begin{itemize}
    \item We study the performance of several state-of-the-art compact deep networks for image classification with several biomedical datasets. In addition, we compare the performance of these networks with standard size deep models and study whether there is an architecture that outperforms the rest.
    \item We perform a thorough statistical analysis to compare the performance of manually designed, automatically designed, and quantized compact networks to answer whether transfer learning can be successfully employed with them in the biomedical context. Moreover, we analyse if there are significant differences about these 3 families of compact networks.
    \item We explore several semi-supervised learning methods to improve the performance of compact deep networks, and study whether any of those methods is especially suited to work with compact models.
    \item We develop a library that simplifies the construction of compact models using semi-supervised learning methods. This is, up to the best of our knowledge, the first library for applying semi-supervised learning methods to construct deep compact image classification models.
\end{itemize}


All the experiments, models, and the library for this work are available at \url{https://github.com/adines/SemiCompact}.

\section{Materials and Methods}\label{sec:experiments}

In this section, we present the datasets, the compact architectures, the procedures and tools used for training and evaluating those architectures, and the semi-supervised learning methods employed in this work. We start by introducing the datasets that have been used for our experiments. 

\subsection{Datasets} \label{sec:datasets}
In this work, we propose a benchmark of 10 partially annotated biomedical datasets, described in Table~\ref{tab:datasets}, and evaluate the performance of deep learning models and semi-supervised methods using such a benchmark.

\begin{table}[h]
\centering
\resizebox{\linewidth}{!}{%
{\small

\begin{tabular}{cccc}
 \toprule
  Dataset & Number of Images & Number of  Classes & Description\\
 \midrule
 \rowcolors{0}{white}{black!10!white}
 Blindness~\cite{blindness}& $3662$ & $5$ & Diabetic retinopathy images\\
 Chest X Ray~\cite{kermany} & $2355$ & $2$ & Chest X-Rays images\\
 Fungi~\cite{fungi} & $1204$ & $4$ & Dye decolourisation of fungal strain\\
 HAM 10000~\cite{ham}& $10015$ & $7$ & Dermatoscopic images of skin lesions\\
 ISIC~\cite{isicDataset} & $1500$ & $7$ & Colour images of skin lesions\\
 Kvasir~\cite{kvasirDataset}  & $8000$ & $8$ & Gastrointestinal disease images\\
 Open Sprayer~\cite{sprayer} & $6697$ & $2$ & Dron pictures of broad leaved docks\\
 Plants~\cite{plantsDataset}  & $5500$ & $12$ & Colour images of plants\\
 Retinal OCT~\cite{kermany} & $84484$ & $4$ & Retinal OCT images\\
 Tobacco~\cite{tobacco}  & $3492$ & $10$ & Document images\\
 \bottomrule
\end{tabular}}}
\caption{Description of the biomedical datasets employed in our experiments.}\label{tab:datasets}
\end{table}

For our study, we have split each of the datasets of the benchmark into two different sets: a training set with the $75 \%$ of images and a testing set with the $25 \%$ of the images. In addition, for each dataset we have selected $75$ images per class using them as labelled images and leaving the rest of the training images as unlabelled images to apply the semi-supervised learning methods. The splits used in our experiments and more information about datasets are available in the project webpage.



\subsection{Compact networks}

We have explored a variety of manually designed, automatically designed, and quantized  architectures. Namely, we have studied 4 manually-designed compact networks (MobileNet~\cite{Mobilenetv2}, ResNet-18~\cite{He16},  SqueezeNet~\cite{SqueezeNet}, and SuffleNet~\cite{ShuffleNet}), 3 automatically designed compact networks (FBNet~\cite{FBNet}, MixNet~\cite{MixNet}, and MnasNet~\cite{MnasNet}) and 2 quantized compact networks (ResNet-18 quantized and ResNet-50 quantized). In addition, for our experiments, we have considered three standard size networks that are ResNet-50~\cite{He16}, ResNet-101~\cite{He16}, and EfficientNet-B3~\cite{efficientnet}. We provide a comparison of different features of these networks in Table~\ref{tab:features}. 

From Table~\ref{tab:features}, we can notice that automatically designed networks achieve better results than manually designed architectures when tested on the ImageNet challenge; and quantized networks obtain similar results to their non-quantized counterparts (a 0.1\% difference in Top-1 accuracy). In addition, it is also worth noting that the accuracy of compact models is getting closer to bigger models, but using less parameters and requiring less floating-point operations (FLOPs).

\begin{table}[h]
    \centering
    \resizebox{\linewidth}{!}{%
    \begin{tabular}{cccccc}
    \toprule
         Name & Params (M) & FLOPs (M) & Top-1 acc (\%)  & Top-5 acc (\%) & Design\\
         \midrule
         ResNet-50 & 26 & 4100 & 76.0 & 93.0 & Manual\\
         ResNet-101 & 44 & 8540 & 80.9 & 95.6& Manual\\
         EfficientNet-B3 & 12 & 1800& 81.6& 95.7& Auto\\
         \midrule
         FBNet &  9.4 & 753 & 78.9 & 94.3 & Auto\\
         MixNet & 5 & 360 & 78.9 & 94.2 & Auto\\
         MNasnet & 5.2 & 403 & 75.6 & 92.7 & Auto\\
         MobileNet v2 &3.4 & 300 & 74.7 & 92.5& Manual\\
         ResNet-18 & 11  & 1300 & 69.6 & 89.2 & Manual\\
         SqueezeNet & 1.3 & 833 & 57.5 & 80.3 & Manual\\
         ShuffleNet v2 & 5.3 & 524 & 69.4 & 88.3 & Manual\\
         \midrule
         ResNet-18 quantized & 11 & - & 69.5 & 88.9 & Quantized\\
         ResNet-50 quantized & 26 & - & 75.9 & 92.8 & Quantized\\

         \bottomrule
    \end{tabular}}
        \caption{Features of the architectures employed in this work. We measure the number of parameters (in millions), the FLOPs (in millions), and the Top-1 and Top-5 accuracy for the ImageNet challenge. In addition, we include how these architectures were designed. Quantized networks change the floating point parameters of the standard version for integer parameters, therefore they have the same number of parameters but do not perform floating point operations.}
    \label{tab:features}
\end{table}

\subsection{Training and evaluation procedure}

All the networks used in our experiments are implemented in Pytorch~\cite{pytorch}, and have been trained thanks to the functionality of the FastAI library~\cite{fastai} using a GPU Nvidia RTX 2080 Ti with 11 GB RAM. In addition, the two quantized newtorks were built using the Pytorch quantization API~\footnote{https://pytorch.org/docs/stable/quantization.html}.


In order to train the models, we have used the transfer-learning method presented in~\cite{fastai}. This is a two-stage procedure that starts from a model pretrained in the Imagenet challenge~\cite{ImageNet}, and can be summarised as follows. In the first stage, we replace the head of the model (that is, the layers that give us the classification of the images), with a new head adapted to the number of classes of each particular dataset. Then, we train these new layers (the rest of the layers stay frozen) with the data of each particular dataset for two epochs. In the second stage, we unfreeze the whole model and retrain all the layers of the model with the new data for fifty epochs. In order to find a suitable learning rate for both the first and second stage, we select the learning rate that decreases the loss to the minimum possible value using the algorithm presented in~\cite{lrate}. Moreover, we employ early stopping based on monitoring the accuracy, and data augmentation~\cite{Simard03} (using flips, rotations, zooms and lighting transformations) to prevent overfitting.

\subsection{Semi-supervised methods}

Now, we describe the semi-supervised learning methods employed in this work. These methods belong to two families of semi-supervised learning techniques: self-training methods~\cite{selftraining} and consistency regularisation techniques~\cite{Sajjadi16}. 

Self-training is a basic approach that (1) defines a base model that is trained on labelled data, (2) uses the model to predict labels for unlabelled data, and, finally, (3) retrains the model with the most confident predictions produced in (2); thus, enlarging the labelled training set. In a variant of self-training called  distillation~\cite{Distillation}, a big model is used for (1) and (2), whereas a faster and smaller model than the model trained in (1) is employed in (3). Data and model distillation are also two forms of self-training. In the case of data distillation~\cite{omni-supervised}, given a model trained on manually labelled data, this technique applies such a model to multiple transformations of unlabelled data, ensembles the multiple predictions, and, finally, retrains the model on the union of manually labelled data and automatically labelled data. In the case of model distillation~\cite{Bucila06}, several models are employed to obtain predictions of unlabelled data; subsequently, those predictions are ensembled, and used to train a new model. Both techniques can also be combined in a technique called data \& model distillation~\cite{omni-supervised}.

In this work, the network used as base model in the plain distillation approach has been ResNet-50 due to the balance between its performance and its efficiency. This network has been also employed in the data distillation method together with 6 different test time augmentation techniques (namely, horizontal flipping, vertical flipping, horizontal and vertical flipping, blurring, Gaussian blurring, and gamma correction). The model distillation method combines the predictions of 3 base networks: ResNet-50, ResNet-101, and EfficientNet-B3. Finally, the model \& data distillation method combines the settings of data distillation and model distillation. All the models were trained using the approach presented in the previous section, and a threshold of $0.8$ was used to select robust annotations from the unlabelled data.

The second family of semi-supervised learning methods studied in this work are consistency regularisation techniques~\cite{Sajjadi16}. This family of methods applies data augmentation to semi-supervised learning based on the idea that a classifier should output the same class distribution for an unlabelled example even after it has augmented. In our case, we have focused on two concrete algorithms: FixMatch~\cite{FixMatch} and MixMatch~\cite{MixMatch}. As a first step, the FixMatch algorithm generates pseudo-labels using the model's predictions on weakly-augmented unlabelled images (that is, images that are flipped and translated); subsequently, for a given image, the prediction is only retained if the model produces a high-confidence prediction; and, finally, the model is trained to predict the pseudo-label when feeding a strongly-augmented version of the same image (that is, an image produced using the RandAugment method~\cite{RandAugment}). Similarly, the MixMatch algorithm uses weakly-augmented images both for generating pseudo-labels and for training the model. For the FixMatch and MixMatch algorithms, we have used the default parameters fixed in the two extensions of FastAI that implement them~\footnote{\url{https://github.com/oguiza/fastai\_extensions}}~\footnote{\url{https://github.com/phanav/fixmatch-fastai}}.

In order to facilitate the reproducibility of our methods, and also to simplify the application of the aforementioned semi-supervised algorithms to image classification problems, we have designed a Python library, available at the project webpage, that implements them. The library provides an API that is summarised in Figure~\ref{fig:pseudocode}. Several settings can be configured for the methods of the API, and we explain those options in the documentation of the project webpage. In order to employ, for instance, the data distillation method, the user must provide the path to the dataset of images (that should contain labelled and unlabelled images), the name of the base model, the name of the final model (both the compact networks and standard size models studied in this work are supported by the library, and new models can be easily included), the list of transformations that will be applied, and the confidence threshold. From that information, the library will automatically train the base model using the labelled images (the model will be trained following the approach presented in the previous section), use such a base model to annotate the unlabelled images, and train the final model (again using the procedure previously presented). The process is similar, and also automatic, for the rest of the methods provided in the library.

\begin{figure}[!ht]
\begin{lstlisting}[frame=single]
plainDistillation(baseModel, 
    targetModel, path, pathUnlabelled, 
    outputPath,confidence)

dataDistillation(baseModel, 
    targetModel, transforms, path, 
    pathUnlabelled, outputPath, 
    confidence)

modelDistillation(baseModels, 
    targetModel, path, pathUnlabelled, 
    outputPath, confidence)

modelDataDistillation(baseModels, 
    targetModel, transforms, path, 
    pathUnlabelled, outputPath, 
    confidence)

FixMatch(targetModel, path, 
    pathUnlabelled, outputPath)

MixMatch(targetModel, path, 
    pathUnlabelled, outputPath)


\end{lstlisting}
\caption{\textcolor{black}{API of the semi-supervised methods provided in our library. The \texttt{BaseModel} is the name of the model used as base model in the distillation procedure. \texttt{BaseModels} is a list of models used as base models in the distillation procedure. The \texttt{TargetModel} is the name of the target model trained with the distillation procedure. The \texttt{Path} parameter refers to the path to the labelled images. The \texttt{PathUnlabelled} is the path to the unlabelled images. The \texttt{OutputPath} is the path where the target model will be saved. The \texttt{Transforms} refers to a list of transformations to apply Data distillation. The \texttt{Confidence} parameter refers to the confidence threshold.}}\label{fig:pseudocode}
\end{figure}


\subsection{Statistical analysis}

Finally, we explain the evaluation procedure applied to analyse the compact models and semi-supervised learning methods. We have selected the F1-score metric to evaluate the performance of the different networks and methods. In particular, we have studied the F1-score for all  binary classification datasets and its macro version for the multi class datasets. The F1-score provides a trade-off between precision and recall, hence we have used this metric to determine whether the results obtained are statistically significant by performing several null hypothesis tests using the methodology presented in~\cite{Garcia10,Sheskin11}. In order to choose between a parametric or a non-parametric test to compare the models, we check three conditions: independence, normality, and heteroscedasticity --- the use of a parametric test is only appropriate when the three conditions are satisfied~\cite{Garcia10}.

The independence condition is fulfilled in our study since the models and methods are tested on independent datasets. We use the Shapiro-Wilk test to check 
normality  --- with the null hypothesis being that the data follow a normal distribution --- and, a Levene test to check heteroscedasticity --- with the null hypothesis being that the results are heteroscedastic. Since more than two models are involved in all the conducted tests, we employ an ANOVA test if the parametric conditions are fulfilled, and a Friedman test otherwise. In both cases, the null hypothesis is that all the models have the same performance. Once the test for checking whether a model is statistically better than the others is conducted, a post-hoc procedure is employed to address the multiple hypothesis testing among the different models. A Bonferroni-Dunn post-hoc procedure, in the parametric case or a Holm post-hoc procedure, in the non-parametric case, is used for detecting significance of the multiple comparisons and the $p$ values should be corrected and adjusted~\cite{Garcia10,Sheskin11}. We have performed our experimental analysis with a level of confidence equal to $0.05$. In addition, the size effect has been measured using Cohen's d~\cite{Cohen69}.

\section{Results and Discussion}\label{sec:results}

In this section, we present a thorough analysis for the results obtained by the 9 compact networks and the 6 semi-supervised processes when applied to the datasets introduced in the previous section. In particular, we first compare the performance of the networks when training using only the labelled data. Then, we study the impact of applying the different semi-supervised methods. From the results obtained in our experiments, we can answer some questions that have driven this research.

\begin{table}[h]
\centering
\resizebox{\linewidth}{!}{%
{\small

\begin{tabular}{ccccccccccc|c}
 \toprule
  & Blindness & Chest X Ray & Fungi & HAM 10000 & ISIC & Kvasir & Open Sprayer & Plants & Retinal OCT & Tobacco & Mean(std)\\
 \midrule
 \rowcolors{0}{white}{black!10!white}
ResNet-50 & 59.3 & 89.9 & \textbf{91.0} & \textbf{54.3} & 87.6 & \textbf{89.0} & 91.3 & 84.3 & 97.4 & \textbf{81.8} & \textbf{82.5(13.5)}\\
ResNet-101 & 58.2 & \textbf{90.7} & 86.9 & 52.0 & 84.0 & 83.8 & 95.8 & 84.3 & 96.4 & 80.1 & 81.2(14.1)\\
EfficientNet & 53.6 & 84.1 & 84.7 & 52.8 & 85.0 & 85.4 & \textbf{96.8} & 84.0 & 98.1 & 72.9 & 79.7(14.8)\\
\midrule
FBNet & 57.5 & 87.4 & 89.0 & 47.2 & 85.2 & 88.9 & 95.4 & 81.8 & 94.9 & 73.3 & 80.1(15.3)\\
MixNet & \textbf{61.8} & 89.5 & 89.7 & 46.9 & \textbf{89.9} & 86.8 & 95.5 & \textbf{86.2} & \textbf{98.9} & 76.7 & 82.2(15.3)\\
MNasNet & 56.2 & 89.2 & 90.3 & 55.8 & 81.9 & 84.6 & 95.7 & 82.5 & 97.4 & 75.3  & 80.9(13.9)\\
MobileNet & 52 & 86.9 & 89.0 & 46.7 & 84.1 & 82.1 & 89.1 & 82.9 & 91.0 & 69.4 & 77.3(15.1)\\
ResNet-18 & 56.3 & 90.3 & 94.2 & 53.7 & 86.8 & 84.1 & 91.6 & 80.0 & 97.7 & 77.5 & 81.2(14.4)\\
SqueezeNet & 50.3 & 88.3 & 79.3 & 43.6 & 76.8 & 80.1 & 90.9 & 78.9 & 93.2 & 75.5 & 75.7(15.5)\\
ShuffleNet & 39.5 & 85.7 & 69.9 & 37.6 & 78.9 & 67.0 & 89.6 & 51.9 & 33.9 & 40.7 & 59.5(20.2)\\
\midrule
ResNet-18 quantized & 45.1 & 77.8 & 88.1 & 47.0 & 86.5 & 84.2 & 91.3 & 75.1 & 91.6 & 55.8 & 74.3(17.2)\\ 
ResNet-50 quantized & 48.6 & 77.2 & 83.2 & 42.9 & 78.6 & 81.1 & 85.4 & 77.7 & 91.6 & 69.7 & 73.6(15.0)\\
 \bottomrule
\end{tabular}}}
\caption{Mean (and standard deviation) F1-score for the standard size models (ResNet-50, ResNet-101 and EfficientNet), compact models (FBNet, MixNet, MNasNet, MobileNet, ResNet-18, SqueezeNet, and ShuffleNet) and quantized models (ResNet-18 quantized and ResNet-50 quantized) for the base training method on the 10 biomedical datasets. The best result is highlighted in bold face.}\label{tab:base}
\end{table}

First of all, we have studied which architecture produce the most accurate model for each dataset and whether there are significant differences among them. As we can see in Table~\ref{tab:base}, ResNet-50 obtains the best results in 4 datasets, MixNet (one of the compact architectures) in 4, ResNet-101 in one, and EfficientNet in another one. In addition, as we show in the first row of Table~\ref{tab:stats1-2}, there are not significant differences among standard size models and 4 compact models that are FBNet, MixNet, MnasNet and ResNet-18. 

We have also conducted an analysis of the performance of each architecture trained using the different semi-supervised learning methods --- a summary of the results is presented in Table~\ref{tab:stats1-1}. A first conclusion from such an analysis, see Table~\ref{tab:stats1-2}, is that there are not significant differences among the FBNet, MixNet, MnasNet and ResNet-18 compact networks and the three standard size models when we apply any self-training method. In fact, those compact networks outperform the standard size models when the Plain Distillation and Data Distillation training methods are applied. Similarly, compact networks generally outperform standard size models when consistency regularisation methods are employed.

\begin{table}[h]
\centering
\resizebox{\linewidth}{!}{%
{\small

\begin{tabular}{ccp{10cm}}
 \toprule
  & Test (ANOVA or Friedman) & After post-hoc procedure\\
 \midrule
 \rowcolors{0}{white}{black!10!white}
Base  & $16.0^{***}$ &  ResNet-50 $\simeq$ EfficientNet, FBNet, MixNet, MNasNet, ResNet-101, ResNet-18; ResNet-50 $>$ MobileNet, SqueezeNet, ShuffleNet, ResNet-18 quantized, ResNet-50 quantized; \\

Plain & $14.4^{***}$ & ResNet-18 $\simeq$ EfficientNet, FBNet, MixNet, MNasNet, ResNet-50, ResNet-101; ResNet-18 $>$ MobileNet, SqueezeNet, ShuffleNet, ResNet-18 quantized, ResNet-50 quantized; \\

Data  & $16.9^{***}$ & MixNet $\simeq$ EfficientNet, FBNet, MNasNet, ResNet-50, ResNet-101, ResNet-18; MixNet $>$ MobileNet, SqueezeNet, ShuffleNet, ResNet-18 quantized, ResNet-50 quantized; \\

Model  & $12.1^{***}$ & ResNet-50 $\simeq$ EfficientNet, FBNet, MixNet, MobileNet, ResNet-101, ResNet-18, SqueezeNet; ResNet-50 $>$ MNasNet, ShuffleNet, ResNet-18 quantized, ResNet-50 quantized; \\ 

DataModel & $17.2^{***}$ & ResNet-50 $\simeq$ EfficientNet, FBNet, MixNet, MNasNet,  MobileNet, ResNet-101, ResNet-18; ResNet-50 $>$ SqueezeNet, ShuffleNet, ResNet-18 quantized, ResNet-50 quantized; \\

FixMatch & $11.8^{***}$ & ResNet-18 $\simeq$ MixNet, MNasNet, MobileNet, ResNet-18 quantized, ResNet-50 quantized; ResNet-18 $>$ EfficientNet, FBNet, ResNet-50, ResNet-101, SqueezeNet, ShuffleNet; \\

MixMatch & $2.6^{**}$ & MNasNet $\simeq$ EfficientNet, FBNet, MobileNet, ResNet-18, ResNet-50 quantized; MNasNet $>$ MixNet, ResNet-50, ResNet-101, SqueezeNet, ShuffleNet, ResNet-18 quantized; \\
 \bottomrule
\end{tabular}}}
\caption{Friedman or ANOVA test for the different studied models and applying several semi-supervised methods. $^{***}p < 0.001$; $^{**}p < 0.01$;  $>$: there are significant differences; $\simeq$: there are not significant differences.}\label{tab:stats1-2}
\end{table}

\begin{table}[h]
\centering
\resizebox{\linewidth}{!}{%
{\small

\begin{tabular}{ccccccccccccc}
 \toprule
  & ResNet-50 & ResNet-101 & EfficientNet & FBNet & MixNet & MNasNet & MobileNet & ResNet-18 & SqueezeNet & ShuffleNet & ResNet-18 quantized & ResNet-50 quantized\\
 \midrule
 \rowcolors{0}{white}{black!10!white}
Base & 82.6(13.5) & 81.2(14.1) & 79.7(14.8) & 80.1 (15.3) & 82.2(15.3) & 80.9(13.9) & 77.3(15.1) & 81.2(14.4) & 75.7(15.5) & 59.5(20.2) & 74.3(17.2) & 73.6(15.0) \\

Plain & 83.6(13.4) & 82.9(13.3) & 82.8(13.3) & 82.6(12.8) & 83.8(13.3) & 82.6(14.2) & 79.1(17.6) & 84.2(13.8) & 81.4(14.9) & 57.8(22.7) & 77.0(15.8) & 70.7(21.2) \\

Data & 83.1(13.7) & 83.0(13.2) & 83.1(14.5) & 83.2(12.2) & \textbf{84.7(12.6)} & 83.5(12.7) & 81.4(15.4) & 82.5(15.5) & 81.7(13.7) & 58.7(21.9) & 77.5(13.5) & 74.4(17.7)\\

Model & 83.3(14.3) & 83.0(14.2) & 81.0(15.2) & 80.5(15.0) & 80.8(12.7) & 77.9(14.8) & 81.2(14.9) & 82.9(14.4) & 79.0(16.9) & 56.2(23.9) & 56.2(33.5) & 62.7(26.7) \\ 

DataModel & 83.7(14.0) & 82.7(15.2) & 82.0(15.7) & 80.7(14.6) & 80.8(14.2) & 80.0(13.9) & 80.5(15.9) & 82.8(14.8) & 79.2(16.6) & 57.5(23.1) & 53.1(32.1) & 61.8(26.5) \\

FixMatch & 64.2(20.0) & 40.1(25.2) & 55.4(26.4) & 60.7(22.0) & 74.4(25.1) & 76.2(22.3) & 74.8(22.9) & 81.2(15.0) & 52.0(18.3) & 53.3(21.2) & 74.4(22.9) & 78.4(18.0) \\

MixMatch & 47.1(28.8) & 51.0(23.2) & 64.3(22.6) & 69.3(17.5) & 49.1(28.9) & 79.4(12.9) & 68.1(25.6) & 51.5(36.8) & 57.7(19.4) & 56.4(20.4) & 61.7(25.9) & 64.5(17.5) \\
 \bottomrule
\end{tabular}}}
\caption{Mean (and standard deviation) F1-score for the different studied models and applying several semi-supervised methods. Methods: Base training (Base), Plain Distillation (Plain), Data Distillation (Data), Data Model Distillation (DataModel), FixMatch procedure (FixMatch), and MixMatch procedure (MixMatch). The best result is highlighted in bold face.}\label{tab:stats1-1}
\end{table}

We have also studied the difference in performance among families of compact networks (manually designed, automatically designed, and quantized). From the results presented in Table~\ref{tab:stats2}, we can conclude that automatically designed compact networks obtain the best results with the base training and all the semi-supervised methods, except for FixMatch; and in most cases have a significant difference with manually designed networks. On the contrary, quantized networks, generally have worse performance than NAS and manually designed networks, except when trained with the FixMatch method, that they achieve the best performance.


\begin{table}[h]
\centering
\resizebox{\linewidth}{!}{%
{\small

\begin{tabular}{cccc|cc}
 \toprule
  & NAS & Manual & Quantized & Test (ANOVA or Friedman) & After post-hoc procedure\\
 \midrule
 \rowcolors{0}{white}{black!10!white}
Base & 81.0(14.7) & 73.4(14.6) & 73.9(15.9) & $13.0^{***}$ & NAS $>$ Manual, Quantized;\\

Plain & 83.1(13.2) & 75.6(15.2) & 73.8(17.7) &  $10.9^{***}$ & NAS $>$ Manual, Quantized;\\

Data & 83.8(12.4) & 76.0(15.0) & 72.2(18.1) &  $13.1^{***}$ & NAS $>$ Manual, Quantized;\\

Model & 81.0(12.4) & 74.7(14.7) & 59.5(27.4) & $5.5^{*}$ & NAS $\simeq$ Manual; NAS$>$ Quantized;\\ 

DataModel &  80.5(14.0) & 75.0(15.1) & 57.4(25.8) & $7.5^{**}$ & NAS $\simeq$ Manual; NAS$>$ Quantized;\\

FixMatch &  70.5(21.8) & 65.3(17.0) & 76.4(19.8) & $6.1^{**}$ & Quantized $\simeq$ NAS; Quantized $>$ Manual; \\

MixMatch &  66.0(17.8) & 58.4(15.3)  & 63.1(21.5) & $1.4^{0.26}$ & NAS $\simeq$ Manual, Quantized; \\
 \bottomrule
\end{tabular}}}
\caption{Mean (and standard deviation) F1-score for the different studied families of models and applying several semi-supervised methods. $^{***}p < 0.001$; $^{**}p < 0.01$; $^{*}p < 0.05$; $>$: there are significant differences; $\simeq$: there are not significant differences.}\label{tab:stats2}
\end{table}









In addition, we have also explored which semi-supervised learning method obtains the best results for each network, see Tables~\ref{tab:stats4-1} and \ref{tab:stats4-2}. For the three standard size models, the self-training methods obtain a similar result to the base training approach and outperform the consistency regularisation methods. In particular, the Data Model Distillation approach achieves the best results and has a significant difference with the FixMatch and MixMatch methods, but not with the base training method. For the automatically designed compact networks, the Data Distillation method outperforms the rest; and, as in the previous case, there are not significant differences with the base training method. For manually designed compact networks, there is not a single method that stands out, however, there is a significant difference between the semi-supervised winner method and the base method. Finally, for quantized networks, no method outperforms the rest.

\begin{table}[h]
\centering
\resizebox{\linewidth}{!}{%
{\small

\begin{tabular}{cccccccc}
 \toprule
  & Base & Plain & Data & Model & DataModel & FixMatch & MixMatch\\
 \midrule
 \rowcolors{0}{white}{black!10!white}
ResNet-50 & 82.6(13.5) & 83.6(13.4) & 83.1(13.7) & 83.3(14.3) & 83.7(14.0) & 64.2(20.0) & 47.1(28.8) \\

ResNet-101 & 81.2(14.1) & 82.9(13.3) & 83.0(13.2) & 83.0(14.2) & 82.7(15.2) & 40.3(25.0) & 51.0(23.2) \\

EfficientNet & 79.7(14.8) & 82.8(13.3) & 83.2(14.5) & 81.0(15.2) & 82.0(15.7) & 55.4(26.4) & 64.3(22.6) \\

FBNet & 80.1(15.3) & 82.6(12.8) & 83.2(12.2) & 80.5(15.0) & 80.7(14.6) & 60.7(22.0) & 69.3(17.5) \\

MixNet & 82.2(15.3) & 83.8(13.3) & 84.7(12.6) & 80.8(12.7) & 80.8(12.7) & 74.4(25.1) & 49.1(28.9) \\

MNasNet & 80.9(13.9) & 82.6(14.2) & 83.5(12.7) & 77.9(14.8) & 80.0(13.9) & 76.2(22.3) & 79.4(12.9) \\

MobileNet & 77.3(15.1) & 79.1(17.6) & 81.4(15.4) & 81.2(14.9) & 80.5(15.8) & 74.9(23.0) & 68.1(25.6)\\

ResNet-18 & 81.2(14.4) & 84.2(13.8) & 82.5(15.5) & 82.9(14.4) & 82.8(14.8) & 81.2(15.0) & 51.5(36.8) \\

SqueezeNet & 75.7(15.5) & 81.4(14.9) & 81.4(13.5) & 79.0(16.9) & 79.2(16.6) & 52.0(18.3) & 57.7(19.4) \\

ShuffleNet & 59.5(20.2) & 57.8(22.7) & 58.7(21.9) & 56.2(23.9) & 57.5(23.1) & 53.3(21.2) & 56.4(20.4) \\

ResNet-18 quantized & 74.2(17.2) & 77.0(15.8) & 77.5(13.4) & 56.2(33.5) & 53.0(32.1) & 74.4(22.9) & 61.7(25.9) \\

ResNet-50 quantized & 73.6(15.0) & 70.7(21.2) & 74.4(17.7) & 62.8(26.7) & 61.4(26.8) & 78.4(18.0) & 64.5(17.5) \\
 \bottomrule
\end{tabular}}}
\caption{Mean (and standard deviation) F1-score for the different semi-supervised methods applied to standard size models (ResNet-50, ResNet-101 and EfficientNet), compact models (FBNet, MixNet, MNasNet, MobileNet, ResNet-18, SqueezeNet, and ShuffleNet) and quantized models (ResNet-18 quantized and ResNet-50 quantized).}\label{tab:stats4-1}
\end{table}

\begin{table}[h]
\centering
\resizebox{\linewidth}{!}{%
{\small

\begin{tabular}{ccc}
 \toprule
  & Test (ANOVA or Friedman) & After post-hoc procedure\\
 \midrule
 \rowcolors{0}{white}{black!10!white}
ResNet-50 & $9.75^{***}$ & DataModel $\simeq$ Base, Plain, Data, Model; DataModel $>$ FixMatch, MixMatch;\\

ResNet-101 & $18.81^{***}$ & DataModel $\simeq$ Base, Plain, Data, Model; DataModel $>$ FixMatch, MixMatch;\\

EfficientNet & $17.69^{***}$ & DataModel $\simeq$ Base, Plain, Data, Model; DataModel $>$ FixMatch, MixMatch;\\

FBNet & $13.7^{***}$ & Data $\simeq$ Base, Plain, Model, DataModel; Data $>$ FixMatch, MixMatch;\\

MixNet & $11.2^{***}$ & Data $\simeq$ Base, Plain; Data $>$ Model, DataModel, FixMatch, MixMatch;\\

MNasNet & $2.7^{*}$ & Data $\simeq$ Base, Plain, DataModel, FixMatch, MixMatch; Data $>$ Model;\\

MobileNet & $3.71^{**}$ & Model $\simeq$ Plain, Data, DataModel, FixMatch, MixMatch; Model $>$ Base;\\

ResNet-18 & $6.71^{***}$ & Plain $\simeq$ Data, Model, DataModel; Plain $>$ Base, MixMatch;\\

SqueezeNet & $36.65^{***}$ & Plain $\simeq$ Data, Model, DataModel; Plain $>$ Base, FixMatch, MixMatch;\\

ShuffleNet & $4.31^{**}$ & Base $\simeq$ Plain, Data, Model, DataModel, MixMatch; Base $>$ FixMatch;\\

ResNet-18 quantized & $5.8^{***}$ & Data $\simeq$ Base, Plain, Model, FixMatch; Data $>$ DataModel, MixMatch;\\

ResNet-50 quantized & $4.5^{***}$ & FixMatch $\simeq$ Data; FixMatch $>$ Base, Plain, Model, DataModel, MixMatch;\\
 \bottomrule
\end{tabular}}}
\caption{Friedman or ANOVA test for the different semi-supervised methods applied to the models. $^{***}p < 0.001$; $^{**}p < 0.01$; $^{*}p < 0.05$;  $>$: there are significant differences; $\simeq$: there are not significant differences.}\label{tab:stats4-2}
\end{table}

In our next set of experiments, we analysed whether there is any semi-supervised method that has a better performance than the rest, and, similarly, which semi-supervised method produces the greatest improvement over the base training approach. In Tables~\ref{tab:stats5-1} and~\ref{tab:stats6-1}, we can see that the Data Distillation method obtains the best value in the Friedman's test average ranking. In addition, from the results presented in Tables~\ref{tab:stats5-2} and~\ref{tab:stats6-2}, we can notice that there is a significant difference in the performance between the Data Distillation method and the rest, except for the Plain Distillation method.

\begin{table}[h]
\centering
\resizebox{0.7\linewidth}{!}{%
{\small

\begin{tabular}{ccc}
 \toprule
  Method & F1-Score & Friedman's test average ranking \\
 \midrule
 \rowcolors{0}{white}{black!10!white}
Base & 77.4(14.5) & 4.2\\
Plain & 79.0(14.3) & 5.8\\
Data & 79.7(13.9) & 6.7\\
Model & 75.4(13.1) & 3.8\\
DataModel & 75.4(13.7) & 4\\
FixMatch & 65.4(18.4) & 2.2\\
MixMatch & 60.0(14.6) & 1.3\\
 \bottomrule
\end{tabular}}}
\caption{Average performance of Base training (Base), Plain Distillation (Plain), Data Distillation (Data), Data Model Distillation (DataModel), FixMatch procedure (FixMatch), and MixMatch procedure (MixMatch).}\label{tab:stats5-1}
\end{table}

\begin{table}[h]
\centering
\resizebox{0.8\linewidth}{!}{%
{\small

\begin{tabular}{cccc|c}
 \toprule
  Method & Z value & p value & adjusted p value & Cohen's d \\
 \midrule
 \rowcolors{0}{white}{black!10!white}
Base & 2.6 & $9.6 \times 10^{-3}$ & 0.02 & 0.16\\
Plain & 0.9 & 0.35 & 0.35 & $4.7 \times 10^{-2}$\\
Model & 3.0 & $2.7\times 10^{-3}$ & 0.01 & 0.31\\
DataModel & 2.8 & $5.2 \times 10^{-3}$ & 0.02 & 0.30\\
FixMatch & 4.7 & $3.2 \times 10^{-6}$ & $1.6 \times 10^{-5}$ & 0.83\\
MixMatch & 5.6 & $2.3 \times 10^{-8}$ & $1.4 \times 10^{-7}$ & 1.3\\
 \bottomrule
\end{tabular}}}
\caption{Adjusted p-values with Holm, and Cohen’s d when comparing semi-supervised learning methods. Control method: Data method.}\label{tab:stats5-2}
\end{table}

\begin{table}[h]
\centering
\resizebox{0.8\linewidth}{!}{%
{\small

\begin{tabular}{ccc}
 \toprule
  Method & F1-Score & Friedman's test average ranking \\
 \midrule
 \rowcolors{0}{white}{black!10!white}
Plain & 1.6(1.8) & 4.9\\
Data & 2.4(1.3) & 5.7\\
Model & -2.0(4.0) & 3.3\\
DataModel & -1.9(3.8) & 3.7\\
FixMatch & -11.9(8.1) & 2.1\\
MixMatch & -17.3(5.0) & 1.3\\
 \bottomrule
\end{tabular}}}
\caption{Average increase of performance of Plain Distillation (Plain), Data Distillation (Data), Data Model Distillation (DataModel), FixMatch procedure (FixMatch), and MixMatch procedure (MixMatch) respect to the base method.}\label{tab:stats6-1}
\end{table}

\begin{table}[h]
\centering
\resizebox{0.8\linewidth}{!}{%
{\small

\begin{tabular}{cccc|c}
 \toprule
  Method & Z value & p value & adjusted p value & Cohen's d \\
 \midrule
 \rowcolors{0}{white}{black!10!white}
Plain & 1.0 & 0.34 & 0.34 & $1.8 \times 10^{-2}$\\
Model & 2.9 & $4.1\times 10^{-3}$ & 0.01 & 1.4\\
DataModel & 2.4 & 0.017 & 0.034 & 1.4\\
FixMatch & 4.3 & $1.7 \times 10^{-5}$ & $6.7 \times 10^{-5}$ & 2.4\\
MixMatch & 5.3 & $1.4 \times 10^{-7}$ & $7.2 \times 10^{-7}$ & 5.1\\
 \bottomrule
\end{tabular}}}
\caption{Adjusted p-values with Holm, and Cohen’s d. Control method: Data method.}\label{tab:stats6-2}
\end{table}

Finally, we have studied the efficiency of the different networks, see Table~\ref{tab:tamTime}. In particular, we have analysed the model size, the time that takes each model to complete a training epoch, and, the time that takes each model when applied for inference with a given image. From the results presented in Table~\ref{tab:tamTime}, we can see that there is a great difference in size between the standard size and compact models. Standard size models are all above 100 MB, whereas the majority of compact models are under 100 MB (being ResNet-18 the only exception). In particular, the difference in size between compact and standard size networks ranges from $30 \%$ to $97 \%$, standing out the ShuffleNet network, however the accuracy of this model is usually lower than the rest. We can also notice that the quantized models reduce the size of original models by almost a $90\%$. Another important point when we test the efficiency of a model is the training time per epoch. In our experiments, using a Nvidia RTX 2080 Ti GPU with 11 GB RAM, standard size networks took approximately between 3 and 4 minutes per epoch. On the contrary, compact networks took less than 100 seconds per epoch (being MixNet the only exception). In this aspect, ShuffleNet stands out from the rest, taking only 15 seconds per epoch. Furthermore, quantized networks reduce the training time per epoch around a $25\%$ from its standard version. Finally, we have studied the inference time of each model. We have calculated the time that takes each model to infer the class of an image in a Intel(R) Xeon(R) W-2145 CPU with 3.70GHz, 16 CPUs cores and 32 GB. Here, we notice again the difference between standard size and compact models. While standard size models take between 200 and 300 ms to classify an image, the compact networks take, at maximum, half the time (the exception is MixNet that takes almost the same time than standard models). This is even more prominent in the case of ResNet-18, SqueezeNet, ShuffleNet, ResNet-18 quantized, and ResNet-50 quantized, that take, at least 4 less time than standard size networks. In particular, quantized versions of the models take between $60 \%$ and $80 \%$ less time than their standard size version.

\begin{table}[h]
\centering
\resizebox{0.8\linewidth}{!}{%
{\small

\begin{tabular}{cccc}
 \toprule
   Network & Size (MB) & Train Time (s)  & Inference Time (ms) \\
 \midrule
 \rowcolors{0}{white}{black!10!white}
ResNet-50 & 294 & 151 & 231\\
ResNet-101 & 512 & 263 & 238\\
EfficientNet & 124 & 210 & 301\\
\midrule
FBNet & 42 & 78 & 121\\
MixNet & 67 & 174 & 222\\
MNasNet & 36 & 64 & 84\\
MobileNet & 41 & 70 & 109\\
ResNet-18 & 135 & 51 & 63\\
SqueezeNet & 15 & 52 & 57\\
ShuffleNet & 0.36 & 15 & 14\\
\midrule
ResNet-18 quantized & 11 & 40 & 25\\
ResNet-50 quantized & 23 & 100 & 44\\
 \bottomrule
\end{tabular}}}
\caption{Efficiency of standard size models (ResNet-50, ResNet-101 and EfficientNet), compact models (FBNet, MixNet, MNasNet, MobileNet, ResNet-18, SqueezeNet, and ShuffleNet) and quantized models (ResNet-18 quantized and ResNet-50 quantized). We measure size (in MB), the training time per epoch (in seconds) and the inference time per image (in milliseconds).}\label{tab:tamTime}
\end{table}


\section{Conclusions and further work}\label{sec:conclusions}

In this paper, we have explored how to combine compact networks with semi-supervised learning models. The results show that, with this combination, we can create compact models that are not only as accurate as bigger models, but also faster and lighter. In particular, we have noticed that when training a model without using semi-supervised methods there are compact networks (FBNet, MixNet, MNasNet and ResNet-18) that obtain a similar performance to standard size networks. Also, when we apply semi-supervised learning methods, specifically, the Plain Distillation method and the Data Distillation method, those models outperform the standard size models. In particular, the best results are obtained when we apply Data Distillation to MixNet and Plain Distillation to ResNet-18. Another conclusion that we can draw from our study is that, in general, automatically designed networks obtain better results than manually designed networks and quantized networks, with the exception of ResNet-18 that obtained similar results than automatically designed networks. Regarding the question of which semi-supervised method produce the best results for each king of architecture, we can conclude that the Data Model Distillation method is the best option for standard size networks; that Data Distillation is the best for automatically designed networks; and, in the rest of cases, there is no a general rule, although the Data Distillation approach generally obtains good results. Finally, compact networks outperform standard size networks in efficiency, that is, in size and in speed. In summary, with this study we can conclude that by applying the Data Distillation method to MixNet, or Plain Distillation to ResNet-18 we can obtain models as accurate as standard size model but, also, faster and lighter. Since these rules might not be the best alternative for a particular case, we have developed a library that simplifies the process of training compact image classification models using semi-supervised learning methods. 

In the future, we plan to extend our work to object detection and semantic segmentation tasks, two instrumental tasks in biomedical images. In those problems, deep models employ as backbones architectures like ResNet or EfficientNet, and we plan to study the impact of replacing those backbones with compact models. Another task that remains as further work is the development of a simple approach to integrate compact models into smartphones and edge devices. This is instrumental to facilitate the creation and dissemination of deep learning models for biomedicine. 

\section*{Acknowledgments}
This work was partially supported by Ministerio de Ciencia e Innovación [PID2020-115225RB-I00 / AEI / 10.13039/501100011033], and FPU Grant 16/06903 of the Spanish MEC.

\bibliographystyle{splncs04}
\bibliography{biblio}







\end{document}